\documentclass{article}
\usepackage{ijcai26}

\usepackage{times}
\usepackage{soul}
\usepackage{url}
\usepackage[hidelinks]{hyperref}
\usepackage[utf8]{inputenc}
\usepackage[small]{caption}
\usepackage{graphicx}
\usepackage{amsmath}
\usepackage{amssymb}
\usepackage{amsthm}
\usepackage{booktabs}
\usepackage{algorithm}
\usepackage{algorithmic}
\usepackage[switch]{lineno}
\usepackage{cleveref}
\usepackage{ufdatastudio}

\newcommand{\Fmean}{\mathcal{F}_{\text{mean}}}   % (formerly F_pattern)
\newcommand{\Fshift}{\mathcal{F}_{\text{shift}}} % (formerly F_h)
\newcommand{\Facc}{\mathcal{F}_{\text{acc}}}     % (formerly F_v)

\title{RISE: Interactive Visual Diagnosis of Fairness in Machine Learning Models}

\author{
Zeyao (Ray) Chen$^1$
\and
Christan Grant$^1$
\affiliations
$^1$University of Florida, Gainesville, FL, USA
\emails
\{chenz1, christan\}@ufl.edu
}

\begin{document}

\maketitle

\begin{abstract}
Evaluating fairness under domain shift is challenging because scalar metrics often obscure exactly where and how disparities arise. We introduce \textit{RISE} (Residual Inspection through Sorted Evaluation), an interactive visualization tool that converts sorted residuals into interpretable patterns. By connecting residual curve structures to formal fairness notions, RISE enables localized disparity diagnosis, subgroup comparison across environments, and the detection of hidden fairness issues. Through post-hoc analysis, RISE exposes accuracy-fairness trade-offs that aggregate statistics miss, supporting more informed model selection. 
\end{abstract}

\section{Introduction}
Fairness-aware machine learning has become essential as algorithmic systems are deployed in high-stakes domains~\cite{barocas2019fairness}. While substantial progress has been made in developing algorithms that optimize multiple fairness criteria, practitioners still struggle to interpret and compare models beyond scalar metrics. 
Traditional fairness measures provide aggregate summaries but obscure where and how disparities arise across the prediction distribution~\cite{hardt2016equality}.
This is critical under domain shift, where a model appear 'fair' on average (e.g., low mean difference) but exhibit severe bias against a specific subgroup of high-risk predictions. Current workflows address this by simply computing more scalar metrics, which only creates conflicting signals without revealing the root cause.
This limitation is especially problematic under distribution shift~\cite{gulrajani2021domain}, where fairness violations emerge in specific regions of the prediction while remaining hidden to global metrics. Moreover, fundamental trade-offs exist among desirable fairness notions~\cite{kleinberg2017inherent}.

In practice, evaluating domain generalization models using multiple scalar fairness metrics often leads to conflicting conclusions: a model may satisfy one criterion while violating another. Such ambiguity makes it difficult for practitioners to form a holistic assessment and poses challenges for responsible deployment. Moving beyond these conflicts requires understanding \emph{where} disparities concentrate, \emph{how} they emerge, and \emph{which} subgroups are affected across the full prediction distribution.

However, scalar metrics alone cannot provide this distributional insight. Current practice typically involves computing multiple fairness scores~\cite{mehrabi2021survey}, often with toolkits such as IBM’s AI Fairness 360~\cite{bellamy2018aif360}, and qualitatively weighing conflicting numbers. This workflow offers limited guidance for diagnosing the structure or location of fairness violations across model predictions.

To address this gap, we introduce \emph{RISE} (Residual Inspection through Sorted Evaluation), an interactive system for visual fairness diagnosis. RISE reveals structured error patterns by organizing prediction residuals into interpretable visual forms, enabling environmental drift detection, algorithm comparison, and subgroup analysis. Supported by mathematically grounded indicators ($\Fmean$, $\Fshift$, $\Facc$), RISE transforms abstract fairness metrics into actionable visual evidence for rapid and interpretable bias discovery.

RISE plots sorted signed residuals (y-axis) against their rank (x-axis). Curves closer to y=0 indicate smaller errors (higher accuracy). Fairness issues appear as systematic separation between subgroup curves and their median rulers. RISE also marks two “twin knees” (a left/convex knee and a right/concave knee) that denote regime changes in the residual distribution; mismatched knee locations across groups indicate that error transitions occur at different percentiles or magnitudes for different subpopulations.

\section{Preliminaries}

% RISE provides a visual framework for post-hoc fairness diagnosis. This section formalizes its core components.

\begin{figure*}[t]
    \centering
    \includegraphics[width=1\linewidth]{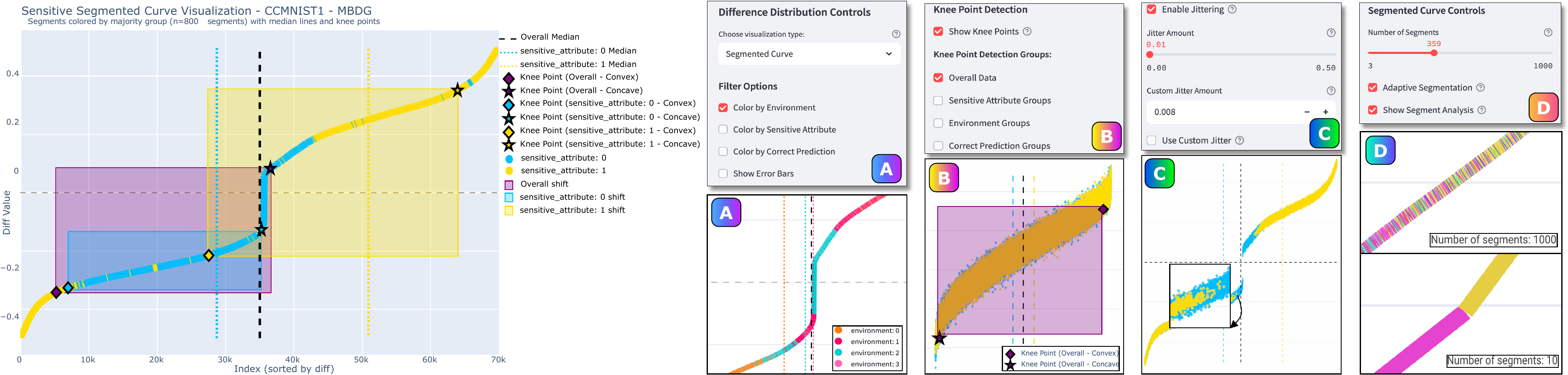}
\caption{
RISE interface on ccMNIST ~\protect\cite{lecun2002gradient} (MBDG shown in~\protect\cite{robey2021model}).
(A) Group coloring for sensitive attributes/environments.
(B) Twin knees (convex $\blacklozenge$, concave $\bigstar$) per group.
(C) Median rulers: overall vs. group alignment (drives $\Fmean$).
(D) Adaptive segmentation reveals local disparities.
Tight, parallel segments indicate good fairness;
near to the x-axis shows higher accuracy.
}

\label{Fig:demo}
\vspace{-4mm}
\end{figure*}

% starting with the signed-error residual, defining the median-based indicators, and introducing the novel fairness metrics derived from them.

% AIF360 and related libraries compute, visualize, and mitigate fairness metrics~\cite{bellamy2018aif360}. RISE goes beyond scalar dashboards by offering a statistical, perception-informed view in which fairness appears as patterns—such as median alignment and knee shifts—on sorted residual curves. This perspective helps address conflicting metrics and reveals how domain shift changes.

\subsection{Positioning and Targeting}
AIF360 and related libraries compute, visualize, and mitigate fairness metrics~\cite{bellamy2018aif360}. Impossibility results show that criteria such as demographic parity, equalized odds, and calibration cannot be simultaneously satisfied~\cite{chouldechova2017fairprediction,kleinberg2017inherent}, leading to conflicting conclusions. Many metrics are also threshold-dependent, producing different fairness assessments at different operating points~\cite{corbett2017measure}. Moreover, aggregate statistics discard distributional structure, masking localized disparities that emerge under domain shift~\cite{sariola2025distributional}. RISE goes beyond scalar metrics by offering a statistical, perception-informed view in which fairness appears as patterns on curves. This perspective helps address conflicting metrics and reveals domain shifts change.

\textsc{RISE} targets two primary audiences. First, ML practitioners deploying models in real-world settings often lack effective tools for diagnosing fairness failures, which are frequently driven by data and distributional factors~\cite{barocas2019fairness,mehrabi2021survey}. Second, \textsc{RISE} supports educational and exploratory use, enabling students and new practitioners to understand fairness mechanisms beyond metrics. By linking fairness indicators to interactive visualizations, \textsc{RISE} makes accuracy–fairness trade-offs  interpretable.

\subsection{Signed-Error Residual Representation}
\label{sec:signed_error_residual}

Let $\mathcal{D}=\{(x_i,y_i,a_i)\}_{i=1}^n$ with $y_i\in\{0,1\}$ and sensitive attribute $a_i\in\{0,1\}$. For a classifier resulting $\hat p_i^{+}=\Pr(y=1\mid x_i)$, we define the signed-error residual
\begin{equation}
d_i=\hat p_i^{+}-y_i\in[-1,1],
\end{equation}
where $d_i>0$ indicates overestimation and $d_i<0$ underestimation.
Let $D=\{d_1,\ldots,d_n\}$ and $D_0=\{d_i\mid a_i=0\},\; D_1=\{d_i\mid a_i=1\}$. With $\operatorname{Med}(\cdot)$ denoting the median of a sorted multiset,
$m_g=\operatorname{Med}(D),\; m_0=\operatorname{Med}(D_0),\; m_1=\operatorname{Med}(D_1),$ and $g$ as entire group.
We normalize median positions to $[0,1]$ via $\bar{m}_g = \frac{\operatorname{Med}(D)}{|D|}$, $\bar{m}_0 = \frac{\operatorname{Med}(D_0)}{|D|}$, and $\bar{m}_1 = \frac{\operatorname{Med}(D_1)}{|D|}$, and re-center group medians by
$\tilde{m}_0=m_0-m_g, \; \tilde{m}_1=m_1-m_g$.
We also normalize x-axis ranks to $[0,1]$ to compare median alignment across groups of different sizes.

\subsection{Core Visual and Analytical Components}

% \textbf{Sorted Residual Visualization.}
RISE sorts residuals $\{d_i\}$ in ascending order, forming a single residual curve.
\textbf{Median Line Analysis.}
Fairness is assessed by comparing group error distributions~\cite{hardt2016equality}. When $m_0\approx m_1\approx m_g$, residuals are balanced; misalignment indicates group-level bias.
\textbf{Understanding the ``Twin Knees.''}
Sorted residual curves are typically S-shaped: values cluster near $0$ and increase in the tails. We define knees as inflection points separating low- and high-error regimes. Earlier or sharper knees for a sensitive group indicate elevated errors over a larger population share.
\textbf{Multi-Group Knee Analysis.}
Using adaptive Kneedle~\cite{satopaa2011kneedle}, we detect a convex left knee ($\ell$) and concave right knee ($r$). For the global set $D$, the knee coordinates are $(d^g_{\ell}, \bar{h}^g_{\ell})$ and $(d^g_{r}, \bar{h}^g_{r})$, where $d$ is the residual and $\bar{h}$ the normalized percentile.
We apply the same procedure to each subgroup $D_a$ ($a\in\{0,1\}$, extensible). For comparison, subgroup knees are mapped into the global sorted-residual space, yielding $(d^a_{\ell}, \bar{h}^a_{\ell})$ and $(d^a_{r}, \bar{h}^a_{r})$ (e.g., $\bar{h}^1_{\ell}$ is group~1’s left-knee percentile relative to $|D|$).

\subsection{Residual-Based Indicators}

From the residual curve, we derive three indicators.

\textbf{\texorpdfstring{$\Fmean$ (Median Alignment Fairness)}{F-pattern}}
$\Fmean$ measures group median alignment: values near $1$ indicate balanced residual distributions, while smaller values suggest systematic bias.
\begin{equation}
\Fmean = 1 - \frac{|\tilde{m}_0 - \tilde{m}_1|}{2}
\end{equation}

For each knee, we compute sensitive--non-sensitive displacements:
\emph{Vertical displacement:}
$
\Delta d_\ell = d^1_{\ell} - d^0_{\ell}, \quad
\Delta d_r = d^1_{r} - d^0_{r}
$
\emph{Horizontal displacement:}
$
\Delta \bar{h}_\ell = \bar{h}^1_{\ell} - \bar{h}^0_{\ell}, \quad
\Delta \bar{h}_r = \bar{h}^1_{r} - \bar{h}^0_{r}
$

We normalize by the corresponding global knee coordinates to obtain relative displacements:
$
\text{V}_\ell = \frac{\Delta d_\ell}{d_{g,\ell}}, \quad
\text{V}_r = \frac{\Delta d_r}{d_{g,r}}, \quad
\text{H}_\ell = \frac{\Delta \bar{h}_\ell}{\bar{h}_{g,\ell}}, \quad
\text{H}_r = \frac{\Delta \bar{h}_r}{\bar{h}_{g,r}}
$

\textbf{\texorpdfstring{$\Fshift$ (Horizontal Knee)}{F-h}}
$\Fshift$ is the mean relative horizontal deviation:
\begin{align}
\Fshift
&= \frac{1}{2}\left|\frac{\bar{h}^1_{\ell}-\bar{h}^0_{\ell}}{\bar{h}_{g,\ell}}\right|
 + \frac{1}{2}\left|\frac{\bar{h}^1_{r}-\bar{h}^0_{r}}{\bar{h}_{g,r}}\right|
= \frac{1}{2}|\text{H}_\ell| + \frac{1}{2}|\text{H}_r|
\end{align}
\textbf{\texorpdfstring{$\Facc$ (Vertical Knee)}{F-v}}
$\Facc$ is the mean relative vertical deviation:
\begin{align}
\Facc
&= \frac{1}{2}\left|\frac{d^1_{\ell}-d^0_{\ell}}{d_{g,\ell}}\right|
 + \frac{1}{2}\left|\frac{d^1_{r}-d^0_{r}}{d_{g,r}}\right|
= \frac{1}{2}|\text{V}_\ell| + \frac{1}{{2}}|\text{V}_r|
\end{align}
Lower values ($\downarrow$) of $\Fshift$ and $\Facc$ indicate that error transitions occur at similar percentiles and magnitudes across groups, implying fewer fairness disparities.

\textsc{RISE} complements these indicators with standard metrics like Accuracy (Acc), Demographic Parity (DP), and Mean Difference (MD) summarized in \Cref{tab:metrics_summary}.

\begin{table}[b]
\centering
\caption{Summary of performance and fairness indicators.}
\label{tab:metrics_summary}
\setlength{\tabcolsep}{4pt}
\footnotesize
\begin{tabular}{@{}llcl@{}}
\toprule
\textbf{Symbol} & \textbf{Brief Definition} & \textbf{Range} & \textbf{Preferred} \\
\midrule
$\mathrm{Acc}$ & Overall accuracy & $[0,1]$ & Higher $\uparrow$ \\
$\mathrm{DP}$ & Positive-rate ratio ($s/n$) & $(0,\infty)$ & Closer to $1$ \\
$\mathrm{MD}$ & Absolute outcome gap & $[0,1]$ & Closer to $0$ \\
\midrule
$\Fmean$  & Group median alignment & $[0,1]$ & Higher $\uparrow$ \\
$\Fshift$ & Knee percentile disparity (H) & $\ge 0$ & Smaller $\downarrow$ \\
$\Facc$   & Knee residual disparity (V)    & $\ge 0$ & Smaller $\downarrow$ \\
\bottomrule
\end{tabular}
\vspace{-5mm}
\end{table}

\section{Demonstration Scenarios}

\begin{figure*}[t]
\centering
\includegraphics[width=\linewidth]{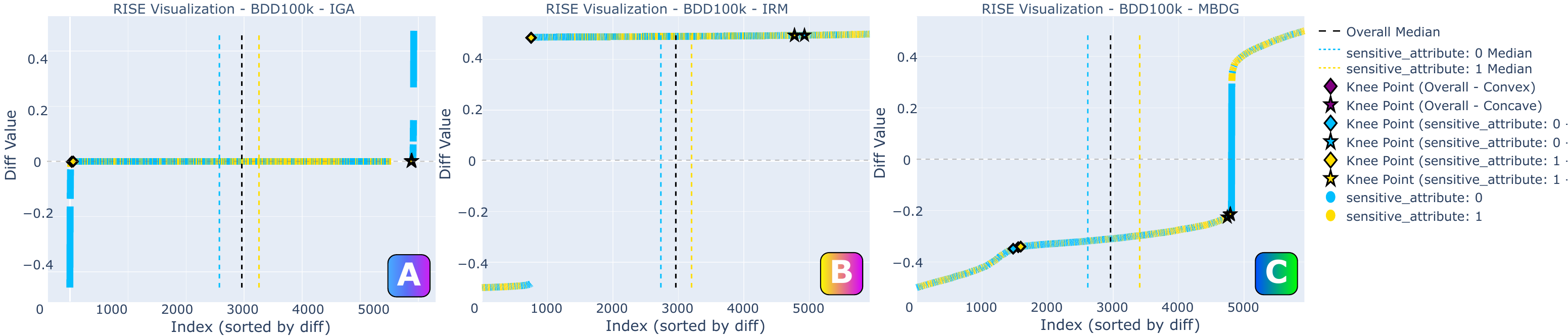}
\caption{Visual signatures of different algorithms on the BDD100K dataset as revealed by RISE. Each plot illustrates the model's accuracy-fairness trade-off. (A) \textbf{IGA}: The curve is close to the x-axis (high accuracy) with minor separation between group, indicating a balanced trade-off. (B) \textbf{IRM}: The tightly clustered, curve far from the x-axis indicates consistent errors across groups (high fairness, $\Facc/\Fshift \approx 0$) but poor low accuracy. (C) \textbf{MBDG}: While overall accuracy is highest, the distorted curve signals significant disparities hidden by metrics.}
\label{Fig:scenarios}
\vspace{-5mm}
\end{figure*}

We demonstrate \textsc{RISE} through three analysis scenarios that highlight its ability to diagnose fairness issues in domain generalization models. These scenarios show how interactive visualizations and residual-based metrics ($\Fmean$, $\Fshift$, $\Facc$) reveal insights that are hidden by number metrics.

\textbf{Setup. } We use the BDD100K dataset~\cite{yu2020bdd100k}, a large-scale driving dataset, to simulate a high-stakes domain shift scenario (e.g., clear vs. rainy weather), to compare three representative algorithms: IRM~\cite{arjovsky2019irm} (fairness-oriented): penalizes instability to improve fairness, often sacrificing overall accuracy, 
MBDG~\cite{robey2021model} (accuracy-oriented) Aligns gradients to find a stable middle ground between accuracy and fairness
and IGA ~\cite{koyama2020invariance} (balanced): Prioritizes predictive power, often hiding localized fairness failures. \Cref{table:bdd_selected} reports the corresponding metrics, while \Cref{Fig:scenarios} visualizes the residual patterns revealed by \textsc{RISE}. The demo is backed by pre-computed results from eight algorithms across five datasets. Attendees interact with a pre-loaded set of models and datasets, and can switch (i) dataset/environment, (ii) protected attribute, and (iii) learning algorithm. For each selection, \textsc{RISE} updates the sorted residual view, subgroup overlays, median rulers, and twin-knee markers in real time; standard metrics (Acc/DP/MD) are pre-computed, while residual indicators ($\Fmean/\Fshift/\Facc$) are computed on the fly.

\textbf{What does Fairness look like in RISE?} 
A fair and accurate model produces residual curves that stay close to the $y=0$ line (low error), overlap tightly across sensitive and non-sensitive groups, and exhibit synchronized inflection points (“knees”) with similar percentile locations and magnitudes. Visually, this appears as flat, well-aligned trajectories with minimal separation. In \Cref{Fig:scenarios}A, the IGA model most closely matches this ideal. These properties are captured by three indicators: $\Fmean\in[0,1]$ measures median alignment (higher is better), $\Fshift$ measures disparity in knee locations (lower is better), and $\Facc$ measures disparity in error magnitude at those transitions (lower is better). Tight, nearly horizontal curves indicate balanced fairness, while proximity to $y=0$ reflects higher accuracy.
% Ideally, a fair and accurate model produces a residual curve that is:Flat and Centered: Hugging the $y=0$ line (indicating high accuracy).Tightly Aligned: The curves for the sensitive and non-sensitive groups should perfectly overlap.Synchronized Knees: The error spikes (knees) for both groups should happen at the exact same percentile and magnitude.Visual Example: In \Cref{Fig:scenarios}A, the IGA model comes closest to this ideal, showing a flat trajectory with minimal separation between group lines.
% \textbf{Key Takeaways.}
% $\Fmean\in[0,1]$ captures group alignment around the residual median (higher is better).
% $\Fshift$ measures where (percentiles) fairness transitions occur across groups (lower is better).
% $\Facc$ measures how large the errors are at those transitions (lower is better).
% Visually, tight, nearly horizontal segments (parallel to the $x$-axis) indicate balanced fairness; closeness to the $x$-axis indicates higher accuracy.

\textbf{Visualizing the Accuracy–Fairness Trade-off. }
This scenario makes the accuracy–fairness trade-off tangible. An attendee is given the confusing metrics in \Cref{table:bdd_selected}. Users will use a dropdown to toggle between the three models. With each click, they will see the plot instantly transform (as shown in \Cref{Fig:scenarios}), revealing each model's distinct visual signature and allowing them to make an balanced choice:
 A high-accuracy model such as \textbf{MBDG} produces residual curves closely aligned with the x-axis, indicating strong predictive performance; however, distortions in the curve reveal localized disparities (\Cref{Fig:scenarios}C).
In contrast, a model explicitly optimized for fairness, such as \textbf{IRM}, exhibits near-perfectly aligned residual curves (high $\Fshift$, $\Facc$), yet a substantial vertical offset from the x-axis immediately signals reduced predictive accuracy (\Cref{Fig:scenarios}A).
A balanced model such as \textbf{IGA} achieves an intermediate configuration, presenting a largely flat curve with minor separations that jointly indicate a favorable balance between fairness and accuracy (\Cref{Fig:scenarios}B).

\begin{table}[t]
\centering
\caption{Performance on selected algorithms (BDD100K)}
\label{table:bdd_selected}
\setlength{\tabcolsep}{4pt} % Adjust column spacing
\footnotesize 
\begin{tabular}{@{}lcccccc@{}}
\toprule
\textbf{Algorithm} & \textbf{Acc} & \textbf{DP} & \textbf{MD} & \textbf{$\Fmean$} & \textbf{$\Fshift$} & \textbf{$\Facc$} \\
\midrule
IGA  & 0.983 & 0.858 & 0.056 & 0.942 & 0.395 & 1.449 \\
IRM  & 0.489 & 0.972 & 0.000 & 0.960 & 0.017 & 0.001 \\
MBDG & 0.960 & 0.828 & 0.023 & 0.933 & 0.042 & 0.040 \\
\bottomrule
\end{tabular}
\vspace{-5mm}
\end{table}

% % ===================== 5. CONCLUSION ========================
\section{Conclusion}
We presented \textsc{RISE}, an interactive system for post-hoc fairness diagnosis under domain shift. By visualizing sorted residuals and coupling them with residual-based indicators ($\Fmean$, $\Fshift$, $\Facc$), \textsc{RISE} enables practitioners to localize disparities, compare algorithms, and reason about accuracy–fairness trade-offs beyond conflicting scalar metrics. The demo illustrates how \textsc{RISE} supports actionable model analysis, training, and selection.
Looking ahead, we plan to extend \textsc{RISE} to multiclass classification and to LLM-based systems. This includes designing for language tasks (e.g., token- or span-level errors, confidence gaps, calibration residuals) and supporting structured subgroup definitions such as text classification, broadening \textsc{RISE} into a general diagnostic interface across modalities.

\bibliographystyle{named}
\bibliography{references}

\end{document}